\documentclass[sigconf,pbalance]{acmart}
\AtBeginDocument{%
  }

\setcopyright{cc}
\setcctype{by}
\copyrightyear{2026}
\acmYear{2026}
\acmDOI{10.1145/3767308.3835904}
\acmConference[MM '26]{Proceedings of the 34th ACM International Conference on Multimedia}{November 10--14, 2026}{Rio de Janeiro, Brazil}
\acmBooktitle{Proceedings of the 34th ACM International Conference on Multimedia (MM '26), November 10--14, 2026, Rio de Janeiro, Brazil}
\acmISBN{979-8-4007-2213-4/2026/11}
\usepackage{enumitem}
\usepackage{colortbl}

\definecolor{top1}{RGB}{135, 206, 250}
\definecolor{top2}{RGB}{224, 241, 255}

\begin{document}
\title{RoES: Rotational Equivariant Selective-frequency Fusion for Multimodal Images}

\author{Jiabao Wang}
\email{hi.wangjiabao@outlook.com}
\affiliation{%
  \institution{Zhongnan University of Economics and Law}
  \city{Wuhan}
  \state{Hubei}
  \country{China}
}

\author{Wenjian Liu}
\email{andylau@cityu.edu.mo}
\affiliation{%
  \institution{City University of Macau}
  \city{Macau}
  \country{China}
}

\author{Yaoming Cai}
\authornote{Corresponding author.}
\email{caiyaom@zuel.edu.cn}
\affiliation{%
  \institution{Zhongnan University of Economics and Law}
  \city{Wuhan}
  \state{Hubei}
  \country{China}
}

\author{Gengyu Zhang}
\email{gzhan@uic.edu}
\affiliation{%
  \institution{University of Illinois Chicago}
  \city{Chicago}
  \state{Illinois}
  \country{United States}
}

\author{Boyan Zhao}
\email{boyan.zhao@outlook.com}
\affiliation{%
  \institution{Zhongnan University of Economics and Law}
  \city{Wuhan}
  \state{Hubei}
  \country{China}
}

\author{Zijia Zhang}
\email{zijiazhang@hubu.edu.cn}
\affiliation{%
  \institution{Hubei University}
  \city{Wuhan}
  \state{Hubei}
  \country{China}
}

\author{Yao Ding}
\email{dingyao.88@outlook.com}
\affiliation{%
  \institution{Xi'an High-tech Research Institute}
  \city{Xi'an}
  \state{Shaanxi}
  \country{China}
}

\author{Xiaobo Liu}
\email{xbliu@cug.edu.cn}
\affiliation{%
  \institution{China University of Geosciences (Wuhan)}
  \city{Wuhan}
  \state{Hubei}
  \country{China}
}

\renewcommand{\shortauthors}{Jiabao Wang et al.}

\begin{abstract}
Infrared-visible image fusion facilitates robust multimodal perception by integrating complementary textural nuances from visible sensors with thermal signatures from infrared systems. Due to the task's inherently ill-posed nature, existing methods heavily rely on structural priors but typically enforce rotation equivariance uniformly across all features. Such a holistic approach overlooks a critical distinction where low-frequency shared structures strictly adhere to equivariant constraints while high-frequency modality-specific details require greater flexibility to preserve unique information. To bridge this gap, we propose \textbf{RoES}, a \textbf{Ro}tational \textbf{E}quivariant \textbf{S}elective-frequency fusion network. Instead of employing static decomposition, we introduce a trainable rotation-enhanced updater/predictor module to dynamically decouple low- and high-frequency components. The resulting representations are then processed through a dual-branch fusion module tailored for spectral consistency. Specifically, a rotation-equivariant Mamba is employed to capture long-range structural dependencies in the low-frequency domain, while a polar spectral attention-based Dual-Fourier block refines high-frequency details under explicit low-frequency guidance. Extensive experiments demonstrate that RoES consistently achieves state-of-the-art performance in both fusion quality and downstream object detection, establishing a robust solution for multimodal fusion by reconciling frequency-selective features with equivariant constraints. The source code is available at \url{https://github.com/BryceLosky/RoES-Fusion}.
\end{abstract}




\maketitle

\section{Introduction}
\label{sec:intro}

Robust visual perception in safety-critical applications such as autonomous driving and surveillance demands reliable scene understanding across diverse and adverse environmental conditions \cite{mmf-pami-25, li2026all, li2025umcfuse, li2026awm}. Infrared-visible image fusion is a prominent approach to this end, combining the complementary characteristics of two widely deployed sensor modalities. Visible cameras provide rich spatial textures and fine structural detail, yet degrade severely under low illumination or glare. Infrared sensors offer illumination-invariant thermal detection, yet sacrifice texture resolution and are susceptible to thermal crossover artifacts~\cite{Spatial-Frequency}. Through cross-modal complementarity, fusion produces imagery that is simultaneously detail-preserving and environmentally robust. This unified representation directly benefits downstream tasks such as object detection~\cite{liu2022tardal} and semantic segmentation~\cite{liu2023segmif}, which require both properties to perform reliably in real-world conditions.

\begin{figure}[t]
    \centering
    \includegraphics[width=0.90\linewidth]{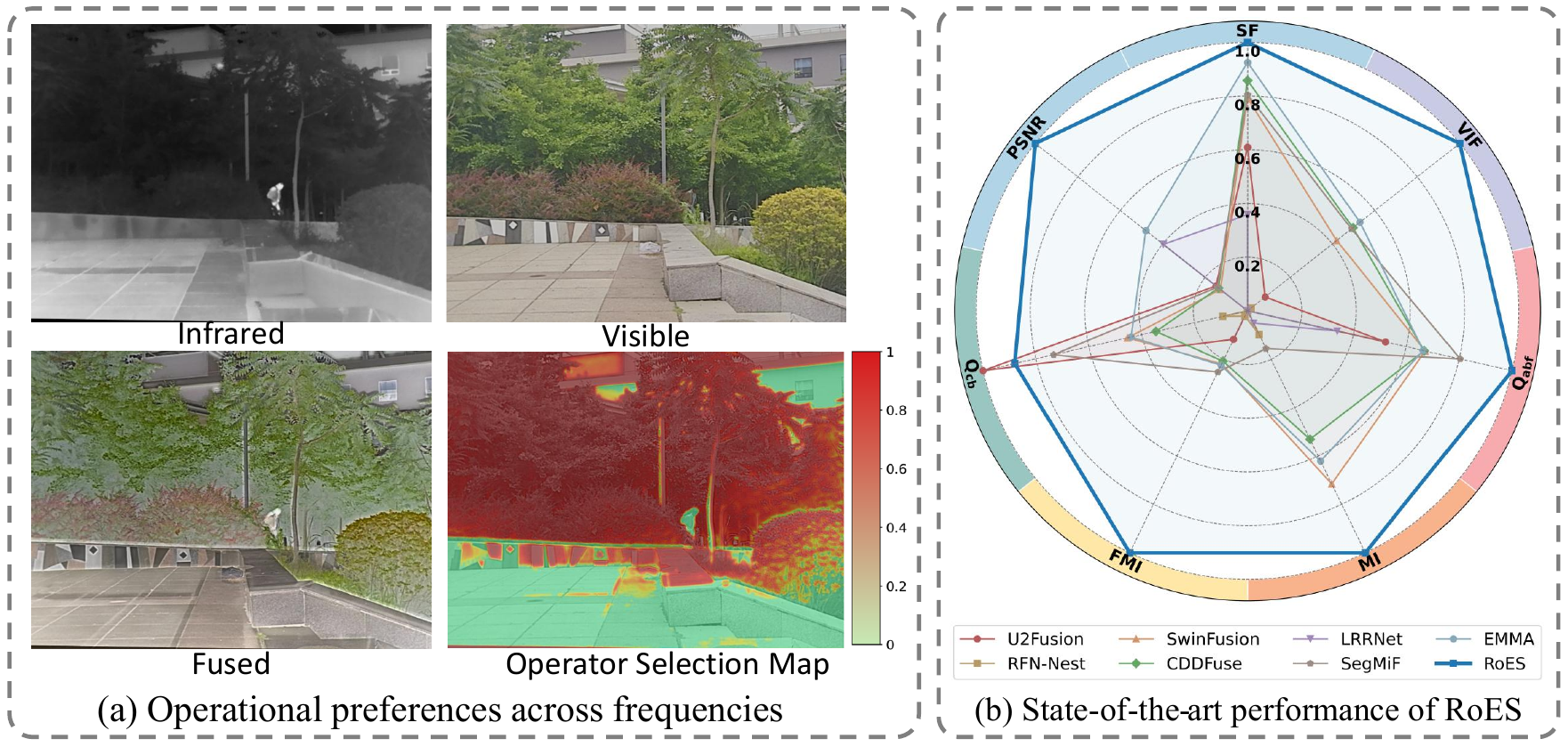}
    \caption{Motivation and performance of RoES. (a) An attention-based router reveals frequency-dependent operator selection in a preliminary U-Net fusion task: rotation-equivariant convolutions for low-frequency structures (green) and standard convolutions for high-frequency textures (red). (b) RoES achieves state-of-the-art performance on the M$^3$FD benchmark.}
    \label{fig:motivation}
\end{figure}

However, infrared-visible image fusion is fundamentally ill-posed. For any registered source pair, no ground-truth image simultaneously preserves all complementary content from both modalities~\cite{mmf-pami-25}. Consequently, fusion models cannot be trained against a definitive target and must instead rely on structural priors to constrain the solution space~\cite{zhang2023visible}. This raises the question of what structural prior to impose.

Among possible priors, geometric consistency---particularly rotation equivariance---has recently emerged as a promising choice \cite{cohen2016gcnn}. It reflects a simple geometric principle: when the input rotates, the representation should transform accordingly in a predictable manner. This inductive bias has shown clear benefits in visual recognition and dense prediction \cite{fconv,weiler2019e2}. More recently, Zhao~\emph{\textit{et al.}}~\cite{EMMA} demonstrated that geometric consistency directly benefits multimodal fusion by introducing rotation-equivariant modeling into the fusion pipeline. 

However, when rotation-equivariant modeling is introduced into fusion, it is typically enforced uniformly across the representation~\cite{EMMA}, overlooking the fact that low- and high-frequency components are not equally compatible with the same geometric prior. As shown in Fig. \ref{fig:motivation}, a fusion model exhibits evident operational preferences across frequency, where standard convolutions are prioritized for  high-frequency textures while rotation-equivariant convolutions are selectively invoked for low-frequency structural components. Applying the same rigid equivariant constraint to both regimes may therefore stabilize shared structure while over-regularizing detail modeling, suppressing the complementary information that fusion is meant to preserve. This concern is also echoed in recent low-level vision studies. Xie \textit{et al.}~\cite{rotation,fconv} note that practical equivariant modules can incur approximation errors on complex local patterns. Bai \textit{et al.}~\cite{bai2025regularization} further show that strict equivariance constraints can limit representation accuracy when the assumed symmetry does not hold precisely. In Sec.~\ref{subsec:method:freq-select}, we formalize this observation as a frequency-selective equivariance constraint and further validate it through a controlled frequency-threshold experiment.

A further challenge is that realizing frequency-selective fusion first requires a meaningful low/high-frequency decomposition, yet many frequency-aware fusion methods still rely on fixed, hand-crafted transforms~\cite{pajares2004wavelet,zhao2023cddfuse,waveMamba} that predefine the boundary between shared structure and modality-specific detail rather than adapting it to the fusion objective.

We therefore propose \textbf{Ro}tational \textbf{E}quivariant \textbf{S}elective-frequency fusion (\textbf{RoES}) to address both the mismatch of uniform equivariance and the rigidity of fixed frequency decomposition. RoES first employs a learnable lifting-based module, RoUP, to adaptively separate multimodal features into low-frequency structure and high-frequency detail. It then selectively imposes rotation equivariance only on the low-frequency branch, where a rotation-equivariant Mamba-style encoder~\cite{eqmamba} produces a geometry-consistent representation with stable global structure. Rather than constraining high-frequency features with the same rigid prior, RoES refines them in the frequency domain through a Frequency-guided Dual-Fourier (FDF) block, where the stable low-frequency representation guides high-frequency spectral optimization via Polar Spectral Attention (PSA). This selective design achieves a better balance between structural robustness and detail preservation, as validated by extensive experiments.

Our contributions are four-fold:
\begin{itemize}[leftmargin=*, itemsep=2pt, topsep=4pt]
    \item \textbf{Frequency-selective equivariance.} We formulate a frequency-selective prior for multimodal fusion, showing that rotation equivariance is more suitable for low-frequency structure than for high-frequency details.
    
    \item \textbf{Learnable decomposition.} We introduce RoUP, a learnable lifting module with an asymmetric CNN/F-Conv predictor/updater and a direction-aware loss for task-adaptive low/high-frequency decomposition.
    
    \item \textbf{RoES framework.} We propose a selective-frequency fusion architecture that uses rotation-equivariant Mamba for low-frequency global structure and a low-frequency-guided Dual-Fourier block for high-frequency refinement.
    
    \item \textbf{Polar spectral refinement.} We design PSA to inject rotation-consistent low-frequency cues into the high-frequency amplitude spectrum in polar coordinates, enabling low-frequency-guided high-frequency refinement in the frequency domain while preserving phase-sensitive localization.
\end{itemize}

\begin{figure*}[tbh!]
    \centering
    \includegraphics[width=1\linewidth]{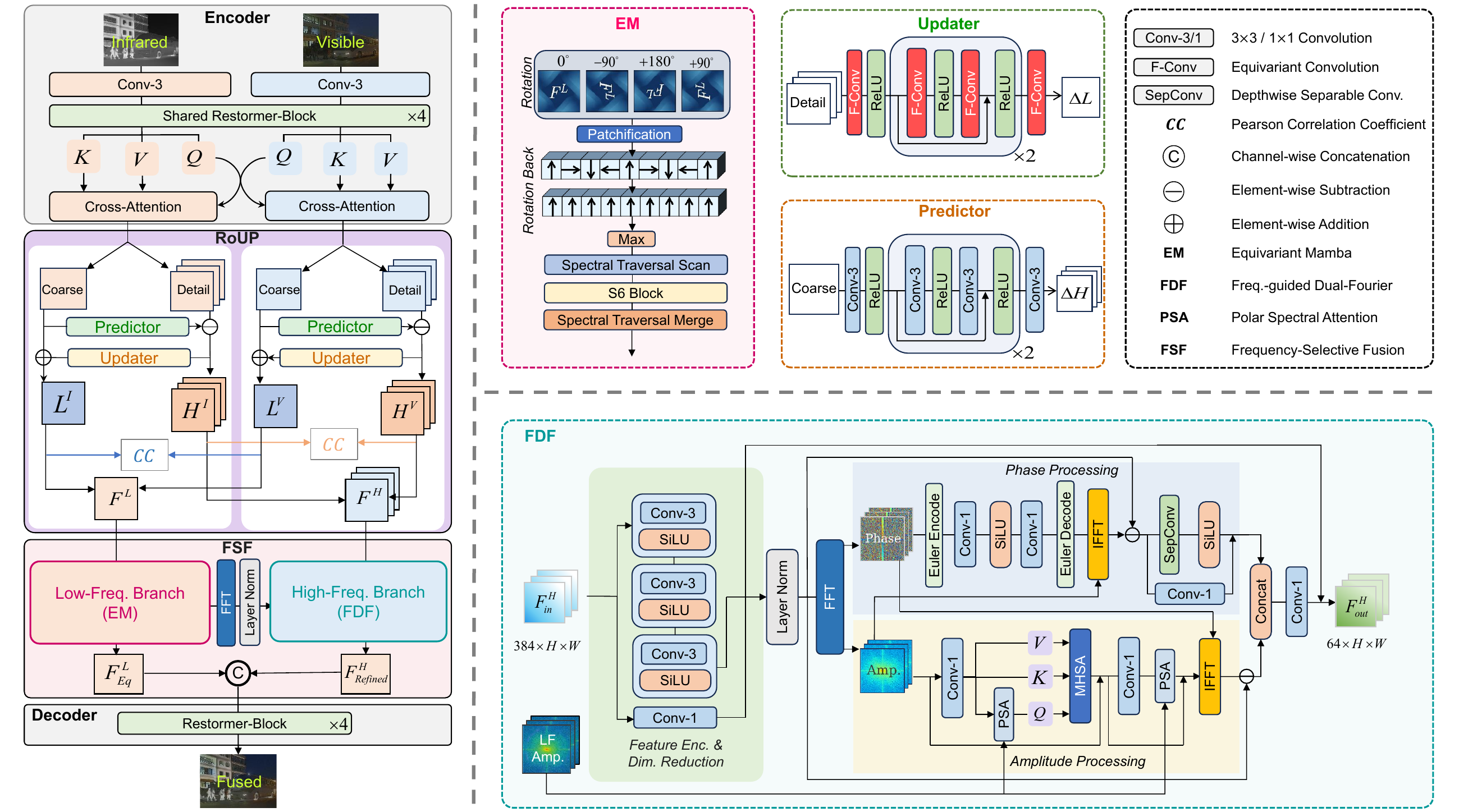}
    \caption{Overview of the proposed RoES model. Paired infrared-visible images are encoded by a shared Restormer encoder and decomposed into low- and high-frequency components via the trainable RoUP module. Our frequency-selective fusion module comprises a rotation-equivariant Mamba block for low-frequency structure modeling, and a frequency-guided Dual-Fourier block for high-frequency detail refinement under low-frequency guidance.}
    \label{fig:overview}
      \vspace{-8pt}
\end{figure*}

\section{Related Work}
\subsection{Infrared-visible Image Fusion}
Infrared-visible image fusion aims to extract and integrate complementary multimodal cues \cite{ref1, ref2, ref3, 10520930, 10879065}. Early methods mainly relied on multi-scale transforms \cite{ref4}, whereas later deep models shifted toward data-driven decomposition \cite{ref5}. Autoencoder- and CNN-based frameworks commonly decompose inputs into base/detail or shared/specific representations \cite{li2019densefuse, ref7, ref8, ref9}, while more recent dual-branch or disentanglement-based designs use correlation-aware objectives to align shared representations and preserve modality-specific textures \cite{ref10, u2fusion, zhao2023cddfuse, ref13, ref14}. Although these strategies improve complementary representation learning, they mostly disentangle information by functional role and rarely consider the different geometric properties of structural low-frequency and detail-dominant high-frequency components during fusion.
Recent studies have also explored Fourier- or spectral-domain interaction for frequency-specific fusion \cite{R1, R3, R4}, but many still rely on fixed transforms, predefined bands, or whole-spectrum processing, limiting adaptive low/high-frequency partitioning and low-to-high structural guidance.
Among learnable decomposition frameworks, the lifting scheme, a second-generation wavelet framework \cite{ref16}, is particularly relevant because it supports task-adaptive decomposition with an explicit inverse path. By factorizing wavelet transforms into split, predict, and update steps \cite{ref17}, it replaces linear filters with learnable nonlinear operators \cite{ref18} while supporting exact or explicit reconstruction under invertible designs \cite{ref15, winnet}. Its effectiveness in image compression, denoising, and time-series modeling \cite{ref19} makes it well suited to adaptive and reversible frequency decoupling in multimodal fusion.

\subsection{Rotation Equivariance Constraint}
For a rotation group $\mathcal{G}$ and $g\in\mathcal{G}$, a mapping $f$ is rotation-equivariant if $f(g\circ\mathbf{X})=g\circ f(\mathbf{X})$, where $g\circ(\cdot)$ denotes the action of $g$ on a feature map. Rotation equivariance thus provides a geometric prior for predictable feature transformations under input rotations \cite{weiler2019e2}. Typical realizations include group convolutions and steerable filters \cite{ref25, ref26}, which are widely used in rotation-sensitive tasks such as oriented object detection and visual matching \cite{ref27, ref29, ref30, ref31, ref32}. For low-level dense prediction, however, strict equivariance is usually realized through finite angular sampling or basis approximations \cite{fconv, pdo}, which can limit accuracy on complex local patterns \cite{rotation, bai2025regularization}. This limitation is particularly relevant to multimodal fusion, where low-frequency components mainly encode stable global structure, whereas high-frequency components are more heterogeneous, modality-dependent, and sensitive to local reliability \cite{rotation_Denoising, bai2025regularization}. Imposing the same rigid equivariant constraint on all frequency bands may therefore be unnecessarily restrictive for detail modeling.
Attention-based fusion offers a flexible mechanism for complementary feature selection \cite{R5, R6, zhao2023cddfuse, liu2025low, LIU2026104498, liu2025selective}, but such geometry-agnostic interaction does not explain how rotation-consistent structure should guide locally variable details. Prior infrared-visible fusion studies have paid limited attention to whether low- and high-frequency branches differ in their compatibility with strict rotation-equivariant modeling, motivating selective rotation priors: explicit equivariant modeling for structurally stable representations and greater flexibility for high-frequency detail fusion.

\section{Methodology}
\label{sec:method}
RoES is built on a simple principle: \emph{rotation equivariance ought to be imposed selectively instead of uniformly across low- and high-frequency components}. As illustrated in Fig. \ref{fig:overview}, we realize this design philosophy through a frequency decomposition and selectivity framework. The detailed architecture and components are elaborated in the following subsections.

\subsection{Frequency-Selective Equivariance Prior}
\label{subsec:method:freq-select}

We treat rotation equivariance as a branch-selective prior at the fusion stage rather than a uniform hard constraint across low- and high-frequency branches, because it is better aligned with low-frequency structural fusion than with high-frequency detail fusion. Let $\mathcal{G}=C_N$ be the discrete rotation group, and let $\mathcal{H}^{\mathcal{G}}\subset\mathcal{H}$ denote the corresponding equivariant subset of the base function class $\mathcal{H}$. Functionally, low-frequency fusion mainly models shared spatial layout, whose response should rotate consistently with the input, making rotation-equivariant modeling suitable. High-frequency fusion instead learns to select reliable modality-specific details, suppress redundant or noisy responses, and enhance complementary edges, textures, or thermal cues under low-frequency guidance. These operations depend on local context and modality reliability rather than rotational symmetry alone, so enforcing the same strict equivariant constraint is unnecessary and can limit detail recovery. RoES therefore applies explicit equivariant modeling only to the low-frequency branch and uses the resulting rotation-consistent low-frequency representation to guide high-frequency fusion. Formally, this branch-selective design is summarized by the following definition.

\begin{definition}[Frequency-selective equivariance constraint]
\label{def:fse}
Let $f_L$ and $f_H$ denote the low- and high-frequency fusion mappings, respectively. A fusion model satisfies the frequency-selective equivariance constraint if
\begin{equation}
f_L \in \mathcal{H}^{\mathcal{G}},\qquad
f_H \in \mathcal{H}.
\end{equation}
That is, the low-frequency branch is constrained to be rotation-equivariant, whereas no strict rotation-equivariant constraint is imposed on the high-frequency branch.
\end{definition}

\noindent\textbf{Controlled validation.}
To examine the scope of Definition~\ref{def:fse} independently of RoES-specific modules, we conduct a controlled experiment with a basic two-branch U-Net. Following FFT decomposition at a frequency threshold $\tau$, components below $\tau$ are processed by a strictly $C_4$-equivariant branch,
$E(\mathbf{x})=\frac{1}{4}\sum_{g\in C_4}g^{-1}B(g\mathbf{x})$,
where $B$ denotes a standard branch, while components above $\tau$ are processed by the standard branch before recombination. Thus, $\tau=0$ applies no equivariance, whereas $\tau=1$ applies it to the full spectrum. The model is trained on MSRS and evaluated on 300 M$^3$FD image pairs.

\begin{figure}[t]
    \centering
    \includegraphics[width=\linewidth]{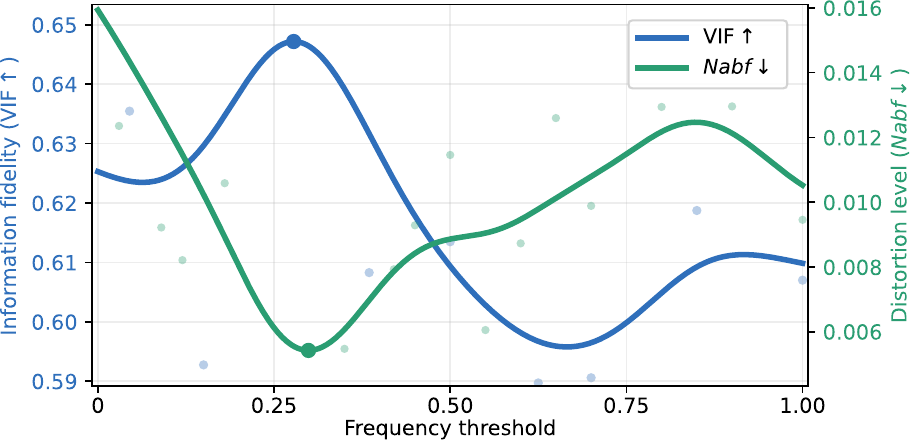}
    \caption{Controlled validation of the frequency-selective equivariance constraint. The threshold $\tau$ spans no equivariance ($\tau=0$) to full-frequency equivariance ($\tau=1$), with both VIF and $N_{abf}$ attaining their optima in the low-frequency regime.}
    \label{fig:frequency-threshold}
\end{figure}

As shown in Fig.~\ref{fig:frequency-threshold}, VIF peaks and $N_{abf}$ reaches its minimum when equivariance is restricted to low frequencies, while extending the constraint toward the full spectrum reverses both gains. VIF measures transferred visual information, whereas $N_{abf}$ quantifies fusion-induced noise and artifacts, jointly covering fidelity preservation and artifact suppression. These aligned optima support the design principle formalized in Definition~\ref{def:fse}, grounding the frequency-selective equivariance constraint in controlled experimental evidence.

In our implementation, $N=4$, i.e., $\mathcal{G}=C_4$. Guided by the above frequency-selective equivariance constraint, RoES first builds interaction-enhanced features with a front-end encoder before RoUP. The encoder contains modality-specific $3\times3$ convolution stems for initial embedding and shallow feature extraction, a parameter-shared Restormer backbone~\cite{restormer} for aligned structural representation learning and long-range context modeling, and a bidirectional cross-attention module for complementary cross-modal information exchange. The resulting encoder features are then directly fed into RoUP for trainable low/high-frequency decomposition.

\subsection{Trainable Frequency Decoupling via RoUP}
\label{subsec:method:RoUP-freq-decoup}

Using the interaction-enhanced encoder features, we employ \emph{Rotation-enhanced Update/Predict networks} (\emph{RoUP}), an asymmetric lifting module with a CNN predictor and an F-Conv updater~\cite{fconv}, to learn a task-adaptive low/high-frequency decomposition for subsequent fusion. Two RoUP modules with the same architecture but independent parameters are used for the infrared and visible streams. For each modality, a wavelet initializer produces initial low/high states, which are then refined by the lifting predictor--updater process.

Consistent with the frequency-selective equivariance constraint (Definition~\ref{def:fse}), RoUP adopts an asymmetric predictor--updater design: a CNN predictor for flexible high-frequency residual estimation and an F-Conv updater for rotation-consistent low-frequency correction. We instantiate the updater with F-Conv~\cite{fconv}, whose Fourier-series filter parametrization improves equivariant filtering accuracy and reduces rotation-induced aliasing. The low- and high-frequency states are then refined through $M$ lifting steps. Specifically, the predictor and updater are shallow residual subnetworks, each comprising an input layer, two residual blocks, and an output layer; the predictor uses standard convolutions, whereas the updater uses F-Conv, with ReLU activations after the intermediate layers. Compared with a fixed wavelet split, this learned refinement is more adaptive while retaining an exact inverse path before fusion.

\subsubsection{Lifting-based decomposition.}
For modality $q\in\{I,V\}$, let $\mathbf{L}^{q,(m)}$ and $\mathbf{H}^{q,(m)}$ denote the low- and high-frequency states at lifting step $m$, respectively. Because the two modalities are processed by independent RoUP instances, their predictors and updaters are modality- and step-specific, yielding the high-frequency residual $\Delta \mathbf{H}^{q,(m)} = \operatorname{Pred}_{q,m}(\mathbf{L}^{q,(m-1)})$ and the low-frequency residual $\Delta \mathbf{L}^{q,(m)} = \operatorname{Upd}_{q,m}(\mathbf{H}^{q,(m)})$. At step $m$, the lifting update is formulated as
\begin{equation}
\mathbf{H}^{q,(m)} = \mathbf{H}^{q,(m-1)} - \Delta \mathbf{H}^{q,(m)}, \qquad
\mathbf{L}^{q,(m)} = \mathbf{L}^{q,(m-1)} + \Delta \mathbf{L}^{q,(m)}.
\end{equation}
Under the lifting interpretation, the CNN-based predictor estimates detail from the coarse state with greater flexibility, so that the high-frequency residual can better preserve modality-specific components, whereas the F-Conv-based updater feeds rotation-consistent structural corrections back to the low-frequency state to progressively emphasize shared structure. After $M$ steps, this yields the modality-specific outputs $\{\mathbf{L}^{I}, \mathbf{L}^{V}, \mathbf{H}^{I}, \mathbf{H}^{V}\}$ for subsequent fusion. Because each lifting step admits an explicit inverse, the decomposition remains exactly reversible before fusion, which helps preserve sparse high-frequency information that is otherwise prone to attenuation during forward propagation.

\subsubsection{Cross-modal decomposition supervision.}
Following the multimodal prior that low-frequency components are more correlated across modalities while high-frequency components are more modality-specific and complementary, we supervise the two parts differently. For the low-frequency part, we compare corresponding structures within matched directional subspaces before aggregation; although the two RoUP modules use independent parameters, subspaces with the same $\theta$ remain geometrically matched because $\theta$ indexes the same F-Conv orientation axis. Compared with global supervision on aggregated features, this subspace-level comparison focuses more directly on within-subspace correspondence and is less sensitive to residual modality-specific differences. For the high-frequency part, redundant cross-modal correlation is suppressed on the final states to promote complementary detail decomposition. Let $\mathbf{L}^{q,\theta}$ denote the low-frequency response of modality $q$ in the $\theta$-th directional subspace produced by the F-Conv-based updater before aggregation, where $\theta\in\{1,\dots,N_\theta\}$, and let $\mathbf{H}^{q}$ denote the final high-frequency state after the lifting steps. Using the Pearson correlation coefficient $\operatorname{CC}\left(\cdot,\cdot\right)$ computed on vectorized feature responses, we define
\begin{equation}
\begin{aligned}
\mathcal{L}_{\mathrm{low}} &=
\frac{1}{N_\theta}\sum_{\theta=1}^{N_\theta}
\left(1-\operatorname{CC}\left(\mathbf{L}^{I,\theta},\mathbf{L}^{V,\theta}\right)\right), \\
\mathcal{L}_{\mathrm{high}} &=
\operatorname{CC}\left(\mathbf{H}^{I},\mathbf{H}^{V}\right)^2.
\end{aligned}
\end{equation}
This supervision encourages agreement of low-frequency shared structure within matched directional subspaces while suppressing redundant high-frequency correlation, thereby promoting complementary detail decomposition in a better-aligned comparison space.

\subsection{Frequency-Selective Fusion}
\label{subsec:method:freq-select-fusion}
After RoUP, the low- and high-frequency components of the two modalities are concatenated as $\mathbf{F}^{L}=\left[\mathbf{L}^{I} \mathbin{\|} \mathbf{L}^{V}\right]$ and $\mathbf{F}^{H}=\left[\mathbf{H}^{I} \mathbin{\|} \mathbf{H}^{V}\right]$, respectively, where $\mathbin{\|}$ denotes channel-wise concatenation. RoES then performs branch-specific fusion on these two components. The low-frequency branch extracts stable shared structure, while the high-frequency branch enhances complementary details under its guidance.

\subsubsection{Equivariant Mamba (EM) Block}
Low-frequency components mainly preserve dominant structure, global layout, and stable spectral energy, and are therefore better suited to global modeling than sparse high-frequency details \cite{waveMamba}. State space models provide global receptive fields with linear complexity, making them effective for large-scale structural modeling. We therefore adopt the spectral state space modeling strategy of Dastani \emph{\textit{et al.}}~\cite{eqmamba} as the low-frequency encoder. In implementation, $\mathbf{F}^{L}$ is first passed through a rotate--Patchification--inverse--max-aggregate pipeline, then through the spectral traversal scan, S6 blocks, and spectral traversal merge for global structural modeling. For the four quarter-turn rotations $r\in C_4$, we rotate $\mathbf{F}^{L}$, apply a shared CNN patchification stem, rotate the responses back by $r^{-1}$, and perform pointwise max aggregation before the spectral SSM stack:
\begin{equation}
\mathbf{F}^{L}_{\mathrm{Eq}}
=
f_{\mathrm{SSM}}\left(
\operatorname{Max}_{r\in C_4}
\left(
r^{-1}\!\circ p\left(r\circ\mathbf{F}^{L}\right)
\right)\right),
\label{eq:low_fusion}
\end{equation}
where $p(\cdot)$ denotes the shared CNN patchification stem, $f_{\mathrm{SSM}}(\cdot)$ denotes the subsequent spectral SSM stack, and $\operatorname{Max}(\cdot)$ denotes pointwise max aggregation. This design injects rotation-consistent structural cues before global spectral modeling and yields the fused low-frequency feature $\mathbf{F}^{L}_{\mathrm{Eq}}$.

\subsubsection{Frequency-guided Dual-Fourier (FDF) Block}
High-frequency details are sparse and more vulnerable to degradation, whereas low-frequency components preserve dominant structure and more stable global statistics \cite{waveMamba}. We therefore use the fused low-frequency feature $\mathbf{F}^{L}_{\mathrm{Eq}}$ to guide high-frequency fusion. We term the proposed high-frequency fusion module the \emph{Frequency-guided Dual-Fourier} block. Since the amplitude spectrum ($\mathbf{A}$) encodes modality-specific features while the phase spectrum ($\boldsymbol{\Phi}$) preserves shared structures \cite{Spatial-Frequency}, our main guidance focuses on amplitude, with only light refinement on phase. Because the preceding wavelet/lifting decomposition yields a $384$-channel concatenated high-frequency representation, we first apply lightweight feature encoding, dimensionality reduction, and layer normalization to obtain $\tilde{\mathbf{F}}^{H}$. We then extract the high-frequency amplitude $\mathbf{A}_H$ and phase $\boldsymbol{\Phi}_H$, and the low-frequency amplitude $\mathbf{A}_L$, via FFT:
\begin{equation}
\operatorname{FFT}(\tilde{\mathbf{F}}^{H}) = \mathbf{A}_H \odot e^{\mathrm{i}\boldsymbol{\Phi}_H}, \qquad 
\mathbf{A}_L = \lvert \operatorname{FFT}(\mathbf{F}^{L}_{\mathrm{Eq}}) \rvert.
\end{equation}

To better exploit low-frequency rotation-consistent structure, the amplitude spectra are mapped to polar coordinates, where rotation-related variations are easier to align along the angular dimension. In this geometry-aligned space, PSA injects low-frequency guidance into the high-frequency amplitude responses; details are given in Sec.~\ref{subsec:method:psa}. Specifically, the first PSA stage produces a low-frequency-guided query for the subsequent multi-head self-attention-based channel recalibration, and the same PSA formulation is reused after channel mixing to further correct the updated amplitude. The corrected amplitude is recombined with the original phase $\boldsymbol{\Phi}_H$ through iFFT to produce a spatial residual. In parallel, the phase path lightly refines $\boldsymbol{\Phi}_H$ in an Euler-encoded representation and recombines it with the original amplitude $\mathbf{A}_H$ through iFFT; the two spatial features are then concatenated, fused by a $1\times1$ convolution, and added to $\tilde{\mathbf{F}}^{H}$ to obtain $\mathbf{F}^{H}_{\mathrm{Refined}}$.

Finally, the fused low- and high-frequency features are concatenated and fed into a decoder of four cascaded Restormer blocks~\cite{restormer} to reconstruct the final fused image. This symmetric decoder helps preserve both global structural consistency and fine local details.

\subsection{Polar Spectral Attention (PSA)}
\label{subsec:method:psa}

\begin{figure}[tbh!]
    \centering
    \includegraphics[width=1\linewidth]{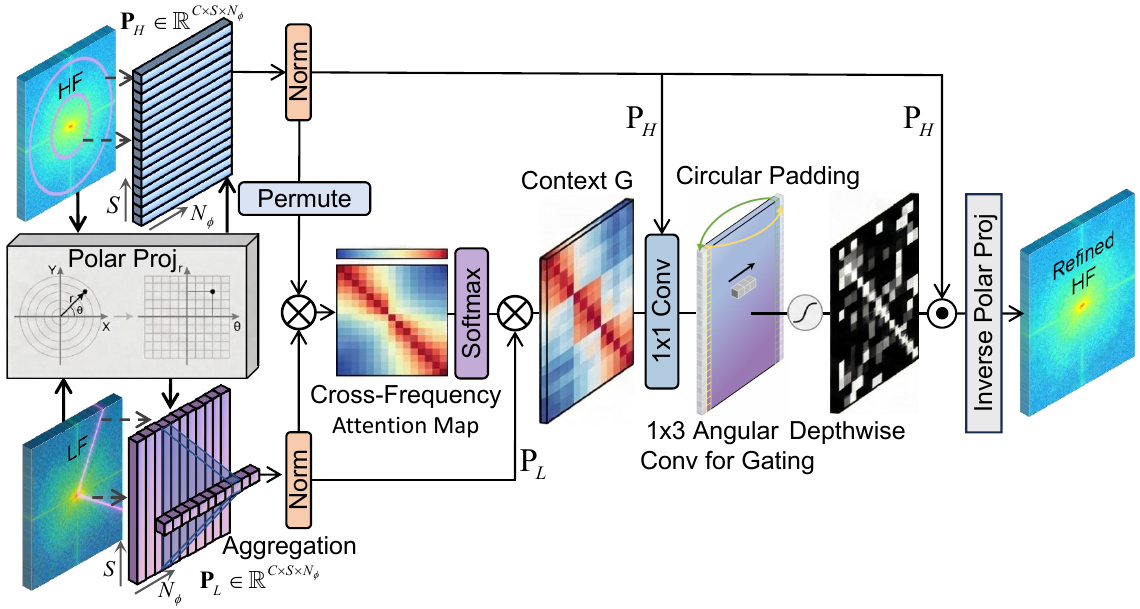}
    \caption{Illustration of polar spectral attention.}
    \label{fig:psa}
    \vspace{-8pt}
\end{figure}

Next, we elaborate on the polar spectral attention module used in the amplitude path of the frequency-guided Dual-Fourier block, depicted in Fig. \ref{fig:psa}. PSA does not rigidly constrain high-frequency features; instead, it injects low-frequency rotation-consistent structure into the high-frequency amplitude spectrum in a geometry-aligned space. This design is motivated by the observation that enhanced low-frequency representations can be used to match and correct high-frequency responses, thereby providing a more stable guidance signal for detail restoration~\cite{waveMamba}. Both PSA stages share the same formulation but use independent parameters.

\subsubsection{Polar Representation}

Let $\mathcal{R}_{\alpha}$ denote rotation by angle $\alpha$, let $A_{\mathbf{x}}=\left|\operatorname{FFT}\left(\mathbf{x}\right)\right|$ be the amplitude spectrum of a feature map $\mathbf{x}$, and let $\Pi(\cdot)$ denote differentiable Cartesian-to-polar resampling. Since the Fourier transform is rotation-equivariant up to coordinate rotation, rotating $\mathbf{x}$ by $\alpha$ rotates its amplitude spectrum by the same angle; in polar coordinates, this becomes a shift along the angular axis:
\begin{equation}
\Pi\!\left(A_{\mathcal{R}_{\alpha}\mathbf{x}}\right)\left(\rho,\phi\right)
=
\Pi\!\left(\mathcal{R}_{\alpha}A_{\mathbf{x}}\right)\left(\rho,\phi\right)
=
\Pi\!\left(A_{\mathbf{x}}\right)\left(\rho,\phi-\alpha\right),
\label{eq:rot_shift}
\end{equation}
where $\rho$ and $\phi$ denote radial frequency and angle, respectively. This converts a geometrically complex rotation in Cartesian spectra into a simpler translation along the angular axis in polar spectra. This conversion is naturally handled by convolution along the angular axis and is also well aligned with similarity computation and softmax-based cross-frequency attention. We therefore project the high- and low-frequency amplitude spectra into polar coordinates:
\begin{equation}
\mathbf{P}_H=\Pi\left(\mathbf{A}_H\right),\qquad
\mathbf{P}_L=\Pi\left(\mathbf{A}_L\right),
\label{eq:polar_proj}
\end{equation}
where $\mathbf{P}_H,\mathbf{P}_L\in\mathbb{R}^{C\times S\times N_\phi}$, $C$ is the number of channels, $S$ is the number of radial bins, and $N_\phi$ is the number of angular bins. Let $\mathbf{P}_H^{\left(u\right)},\mathbf{P}_L^{\left(u\right)}\in\mathbb{R}^{C\times N_\phi}$ denote the $u$-th radial ring. This representation provides a geometry-aligned space for PSA to perform low-to-high structural matching.

\subsubsection{Cross-Frequency Matching and Angular Gating}
We first aggregate the low-frequency polar spectrum into a global descriptor,
\begin{equation}
\bar{\mathbf{P}}_L
=
\operatorname{Norm}_{\phi}\!\left(\sum_{u=1}^{S}\mathbf{P}_L^{(u)}\right),
\end{equation}
where $\operatorname{Norm}_{\phi}$ denotes $L_2$ normalization along the $\phi$ axis.
For each high-frequency ring $u \in \{1, \dots, S\}$, we treat the normalized high-frequency response as the query $\mathbf{Q}^{(u)}=\operatorname{Norm}_{\phi}(\mathbf{P}_H^{(u)})$, and the global low-frequency descriptor as both the key and value $\mathbf{K}=\mathbf{V}=\bar{\mathbf{P}}_L$. The cross-frequency guidance $\mathbf{G}^{(u)}$ for this specific ring is then computed via a scaled dot-product attention:
\begin{equation}
\mathbf{G}^{(u)}
=
\operatorname{Softmax}_{\mathrm{row}}\!\left(
\frac{\mathbf{Q}^{(u)}\mathbf{K}^{\top}}{\sqrt{N_\phi}}
\right)\mathbf{V}.
\label{eq:guidance_ring}
\end{equation}
Stacking $\{\mathbf{G}^{\left(u\right)}\}_{u=1}^{S}$ yields the guidance tensor $\mathbf{G}\in\mathbb{R}^{C\times S\times N_\phi}$. Essentially, each high-frequency radial ring is matched to the global low-frequency descriptor. Because a radial ring corresponds to one spatial frequency across different orientations, the resulting guidance acts as a frequency-aware cross-frequency attention that lets each ring query low-frequency structural cues for detail recovery at its own scale.

The concatenation of $\mathbf{P}_H$ and $\mathbf{G}$ is processed by a $1\times1$ convolution and a circularly padded $1\times3$ angular depthwise convolution to preserve continuity across the $0$--$2\pi$ boundary;  the resulting gate $2\sigma(\cdot)$, where $\sigma(\cdot)$ denotes the sigmoid function, rescales $\mathbf{P}_H$ to produce $\tilde{\mathbf{P}}_H$. Finally, after reverting to Cartesian coordinates, the PSA block outputs the refined high-frequency amplitude:
\begin{equation}
\tilde{\mathbf{A}}_H = \operatorname{PSA}\left(\mathbf{A}_H,\mathbf{A}_L\right).
\label{eq:psa_output}
\end{equation}
In this way, PSA acts as a geometry-aware cross-frequency attention module that aligns low- and high-frequency spectra in polar coordinates and extracts structurally reliable low-frequency cues to guide high-frequency detail recovery.

\subsection{Learning Objectives}

The overall training objective of RoES is formulated as:
\begin{equation}
\mathcal{L}
=
\mathcal{L}_{\mathrm{fuse}}
+
\lambda_{\mathrm{RoUP}}\mathcal{L}_{\mathrm{RoUP}},
\end{equation}
where $\mathcal{L}_{\mathrm{fuse}}$ is the primary image fusion loss, and $\mathcal{L}_{\mathrm{RoUP}}$ is the auxiliary decomposition loss. The $\lambda$ coefficients denote the empirical trade-off weights.

\subsubsection{Fusion objective}
The fusion loss ensures that the final fused image $\mathbf{I}_F$ faithfully integrates complementary information from the infrared ($\mathbf{I}_I$) and visible ($\mathbf{I}_V$) inputs:
\begin{equation}
\mathcal{L}_{\mathrm{fuse}}
=
\mathcal{L}_{\mathrm{int}}
+
\lambda_{\mathrm{grad}}\mathcal{L}_{\mathrm{grad}}
+
\lambda_{\mathrm{ssim}}\mathcal{L}_{\mathrm{ssim}}
+
\lambda_{\mathrm{tv}}\mathcal{L}_{\mathrm{tv}}
+
\lambda_{\mathrm{vgg}}\mathcal{L}_{\mathrm{vgg}},
\end{equation}
where $\mathcal{L}_{\mathrm{int}}$ and $\mathcal{L}_{\mathrm{grad}}$ denote the $L_1$ penalty on pixel intensities and spatial gradients, respectively. $\mathcal{L}_{\mathrm{ssim}}$ enforces structural similarity, $\mathcal{L}_{\mathrm{tv}}$ suppresses local artifacts via total variation, and $\mathcal{L}_{\mathrm{vgg}}$ improves perceptual fidelity.

\subsubsection{RoUP objective}
The RoUP loss ensures accurate frequency decoupling and faithful feature reconstruction. Let $\mathbf{X}_q$ denote the interaction-enhanced encoder feature for modality $q\in\{I,V\}$, and let $\hat{\mathbf{X}}_q = \operatorname{InvRoUP}(\mathbf{L}^{q},\mathbf{H}^{q})$ denote the feature reconstructed by the exact inverse RoUP transform. We first define the reconstruction loss $\mathcal{L}_{\mathrm{rec}}$ to preserve feature fidelity:
\begin{equation}
\mathcal{L}_{\mathrm{rec}}
=
\sum_{q\in\{I,V\}} \Bigl( \|\hat{\mathbf{X}}_q-\mathbf{X}_q\|_{1} + \lambda_{\mathrm{str}}\bigl(1-\operatorname{SSIM}(\hat{\mathbf{X}}_q,\mathbf{X}_q)\bigr) \Bigr),
\end{equation}
where $\lambda_{\mathrm{str}}$ controls the weight of the structural similarity penalty. The overall RoUP objective is then formulated by combining $\mathcal{L}_{\mathrm{rec}}$ with the cross-modal decomposition losses:
\begin{equation}
\mathcal{L}_{\mathrm{RoUP}}
=
\mathcal{L}_{\mathrm{rec}}
+
\lambda_{\mathrm{low}}\mathcal{L}_{\mathrm{low}}
+
\lambda_{\mathrm{high}}\mathcal{L}_{\mathrm{high}},
\end{equation}
where $\mathcal{L}_{\mathrm{low}}$ and $\mathcal{L}_{\mathrm{high}}$ (defined in Sec.~\ref{subsec:method:RoUP-freq-decoup}) supervise the shared low-frequency and modality-specific high-frequency components, respectively.

\section{Experiments}
\subsection{Experimental Setup}
\subsubsection{Datasets}
We train RoES on 1,602 infrared-visible pairs from MSRS~\cite{tang2022piafusion} and evaluate it directly on M$^3$FD~\cite{liu2022tardal}, TNO~\cite{toet2017tno}, and RoadScene~\cite{xu2020fusiondn} without fine-tuning. The differences in scenes, imaging conditions, and data distributions provide a rigorous test of cross-dataset robustness and generalization.

\subsubsection{Comparison Methods}
We compare RoES with the 17 CNN-, GAN-, transformer-, and task- or semantic-aware fusion methods listed in the result tables. We use official implementations and released weights with identical inputs, preprocessing, test splits, and evaluation protocols unless stated otherwise.

\subsubsection{Implementation Details}
RoES is implemented in PyTorch and trained end-to-end on MSRS using one NVIDIA RTX 5090 GPU. Inputs are resized to $128 \times 128$ for training, and AdamW uses an initial learning rate of $5.05\times10^{-5}$ and weight decay of $1\times10^{-4}$. The loss weights are set to $\lambda_{\mathrm{grad}}=26.7$, $\lambda_{\mathrm{ssim}}=25.6$, $\lambda_{\mathrm{tv}}=16.0$, $\lambda_{\mathrm{vgg}}=3.8$, $\lambda_{\mathrm{RoUP}}=33.0$, $\lambda_{\mathrm{str}}=5.6\times10^4$, $\lambda_{\mathrm{low}}=0.003$, and $\lambda_{\mathrm{high}}=0.03$. We use four RoUP lifting steps, a width-64 four-layer spectral SSM, eight attention heads, and two PSA stages. Polar resampling uses $(S,N_\phi)=(512,2048)$ during training and $(3072,3072)$ during inference; the random seed is fixed to 42, and rotation augmentation is disabled.

\subsubsection{Metrics}
We report nine standard metrics: entropy (EN), spatial frequency (SF), average gradient (AG), visual information fidelity (VIF), $Q_{abf}$, mutual information (MI), feature mutual information (FMI), $Q_{cb}$, and peak signal-to-noise ratio (PSNR). EN, MI, and FMI measure retained information; SF and AG measure detail; and the remaining metrics assess fidelity and structural preservation. Higher is better for all metrics. We also compute TOPSIS rankings from normalized scores to summarize overall performance.

\begin{figure}[t]
    \centering
    \resizebox{\linewidth}{!}{%
    \includegraphics[width=1\linewidth]{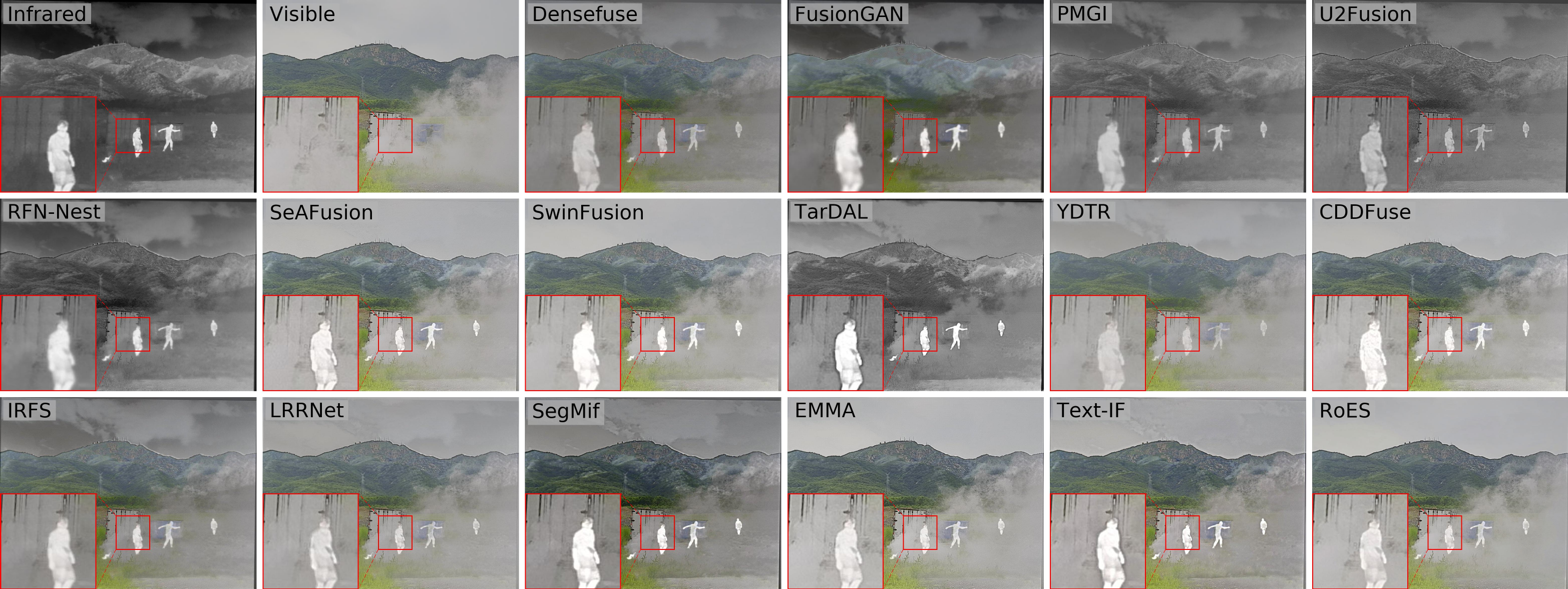}
    }
    \caption{Qualitative comparison of infrared-visible fusion results on the M$^3$FD dataset.}
    \label{fig:fusion-vis-m3df}
    \vspace{-8pt}
\end{figure}

\begin{table}[tbh!]
\centering
\caption{Quantitative evaluation of fusion models on the M$^3$FD, RoadScene, and TNO datasets.}
\resizebox{\linewidth}{!}{%
\begin{tabular}{lcccccccccc}
\toprule
Model & EN$\uparrow$ & SF$\uparrow$ & AG$\uparrow$ & VIF$\uparrow$ & $Q_{abf}$$\uparrow$ & MI$\uparrow$ & FMI$\uparrow$ & $Q_{cb}$$\uparrow$ & PSNR$\uparrow$ & TOPSIS$\uparrow$ \\

\midrule
\multicolumn{11}{c}{\textbf{M$^3$FD Dataset}} \\
\midrule
DenseFuse~\cite{li2019densefuse}  & 6.426 & 9.206 & 3.200 & 0.653 & 0.419 & 2.777 & 0.251 & 0.431 & 15.762 & \cellcolor{top1!17}0.280 \\
FusionGAN~\cite{fusiongan}  & 6.636 & 8.001 & 2.685 & 0.479 & 0.266 & 2.965 & 0.217 & 0.351 & 15.827 & \cellcolor{top1!0}0.067 \\
PMGI~\cite{PMGI}       & 6.304 & 9.199 & 3.228 & 0.588 & 0.368 & 3.011 & 0.226 & 0.425 & 16.117 & \cellcolor{top1!12}0.219 \\
U2Fusion~\cite{u2fusion}   & 6.661 & 12.762 & 4.696 & 0.664 & 0.508 & 2.668 & 0.264 & \cellcolor{top2}0.504 & 15.945 & \cellcolor{top1!34}0.495 \\
RFN-Nest~\cite{ref7}   & 6.864 & 7.827 & 2.869 & 0.644 & 0.362 & 2.812 & 0.247 & 0.436 & 15.564 & \cellcolor{top1!12}0.221 \\
SeAFusion~\cite{SeAFusion}  & 6.848 & 14.476 & 4.923 & 0.737 & 0.549 & 3.340 & 0.275 & 0.449 & 15.631 & \cellcolor{top1!42}0.600 \\
SwinFusion~\cite{swinfusion} & 6.807 & 14.246 & 4.777 & 0.766 & 0.550 & \cellcolor{top2}3.718 & 0.282 & 0.463 & 15.891 & \cellcolor{top1!44}0.630 \\
TarDAL~\cite{liu2022tardal}     & \cellcolor{top1}7.152 & 12.539 & 4.116 & 0.643 & 0.377 & 3.091 & 0.239 & 0.432 & 15.216 & \cellcolor{top1!24}0.374 \\
YDTR~\cite{ydtr}       & 6.547 & 12.001 & 3.908 & 0.678 & 0.457 & 3.007 & 0.255 & 0.416 & 16.509 & \cellcolor{top1!27}0.406 \\
\midrule
CDDFuse~\cite{zhao2023cddfuse}    & 6.905 & 14.782 & 4.844 & 0.789 & 0.549 & 3.447 & 0.280 & 0.455 & 15.914 & \cellcolor{top1!44}0.629 \\
IRFS~\cite{IRFS}       & 6.744 & 10.800 & 3.503 & 0.686 & 0.458 & 2.747 & 0.258 & 0.414 & 15.708 & \cellcolor{top1!23}0.356 \\
LRRNet~\cite{lrrnet}     & 6.435 & 10.734 & 3.584 & 0.639 & 0.457 & 2.743 & 0.243 & 0.429 & 16.582 & \cellcolor{top1!21}0.339 \\
SegMiF~\cite{liu2023segmif}     & \cellcolor{top2}6.985 & 14.347 & 4.828 & 0.787 & 0.587 & 2.897 & 0.288 & 0.484 & 15.926 & \cellcolor{top1!43}0.619 \\
EMMA~\cite{EMMA}       & 6.922 & 15.327 & \cellcolor{top1}5.325 & 0.799 & 0.548 & 3.579 & 0.283 & 0.462 & 16.793 & \cellcolor{top1!47}0.668 \\
Text-IF~\cite{Text-IF}    & 6.931 & \cellcolor{top2}15.755 & \cellcolor{top2}5.284 & \cellcolor{top2}0.852 & \cellcolor{top2}0.611 & 3.351 & 0.301 & \cellcolor{top1}0.505 & \cellcolor{top2}17.145 & \cellcolor{top1!51}0.721 \\
SAGE~\cite{wu2025every} & 6.850 & 13.867 & 4.636 & 0.774 & 0.585 & 3.083 & \cellcolor{top2}0.345 & 0.449 & 15.172 & \cellcolor{top1!46}0.656 \\
DAFusion~\cite{guan2026domain} & 6.706 & 11.347 & 3.951 & 0.736 & 0.564 & 3.021 & 0.344 & 0.456 & 15.528 & \cellcolor{top1!39}0.557 \\
\midrule
RoES (Ours)& 6.858 & \cellcolor{top1}15.932 & 5.098 & \cellcolor{top1}0.941 & \cellcolor{top1}0.642 & \cellcolor{top1}4.135 & \cellcolor{top1}0.421 & 0.495 & \cellcolor{top1}18.127 & \cellcolor{top1!100}0.957 \\

\midrule
\multicolumn{11}{c}{\textbf{RoadScene Dataset}} \\
\midrule
DenseFuse~\cite{li2019densefuse}  & 6.832 & 11.687 & 4.536 & 0.676 & 0.482 & 2.868 & \cellcolor{top2}0.366 & 0.508 & 14.470 & \cellcolor{top1!32}0.496 \\
FusionGAN~\cite{fusiongan}  & 7.065 & 8.468 & 3.310 & 0.457 & 0.252 & 2.730 & 0.245 & 0.450 & 12.645 & \cellcolor{top1!0}0.076 \\
PMGI~\cite{PMGI}       & 6.319 & 11.465 & 4.671 & 0.597 & 0.405 & 3.025 & 0.285 & 0.520 & 14.316 & \cellcolor{top1!24}0.387 \\
U2Fusion~\cite{u2fusion}   & 6.889 & 14.523 & 5.854 & 0.651 & 0.503 & 2.719 & 0.344 & \cellcolor{top2}0.528 & \cellcolor{top2}15.740 & \cellcolor{top1!42}0.617 \\
RFN-Nest~\cite{ref7}   & 7.338 & 7.811 & 3.309 & 0.587 & 0.297 & 2.724 & 0.273 & 0.472 & 13.969 & \cellcolor{top1!9}0.188 \\
SeAFusion~\cite{SeAFusion}  & 7.340 & 18.155 & \cellcolor{top2}7.055 & 0.659 & 0.488 & 3.012 & 0.329 & 0.484 & 15.043 & \cellcolor{top1!51}0.744 \\
SwinFusion~\cite{swinfusion} & 6.990 & 17.314 & 6.357 & 0.696 & 0.438 & 3.254 & 0.327 & 0.494 & 14.299 & \cellcolor{top1!48}0.700 \\
TarDAL~\cite{liu2022tardal}     & 7.320 & 13.560 & 4.768 & 0.604 & 0.415 & \cellcolor{top2}3.270 & 0.316 & 0.443 & 15.041 & \cellcolor{top1!31}0.481 \\
YDTR~\cite{ydtr}       & 6.909 & 13.919 & 5.209 & 0.681 & 0.477 & 2.984 & 0.336 & 0.480 & 15.348 & \cellcolor{top1!38}0.573 \\
\midrule
CDDFuse~\cite{zhao2023cddfuse}    & \cellcolor{top1}7.482 & \cellcolor{top2}18.953 & 6.786 & 0.686 & 0.469 & 3.016 & 0.335 & 0.491 & 14.863 & \cellcolor{top1!52}0.752 \\
IRFS~\cite{IRFS}       & 7.007 & 11.683 & 4.465 & 0.651 & 0.446 & 2.735 & 0.342 & 0.441 & \cellcolor{top1}15.892 & \cellcolor{top1!29}0.450 \\
LRRNet~\cite{lrrnet}     & 7.133 & 12.513 & 4.674 & 0.552 & 0.341 & 2.773 & 0.252 & 0.508 & 12.107 & \cellcolor{top1!19}0.323 \\
SegMiF~\cite{liu2023segmif}     & 7.341 & 15.701 & 6.027 & 0.693 & 0.512 & 2.705 & 0.344 & 0.482 & 15.436 & \cellcolor{top1!45}0.661 \\
EMMA~\cite{EMMA}       & \cellcolor{top1}7.482 & 16.159 & 6.188 & 0.691 & 0.436 & 3.219 & 0.314 & 0.504 & 14.805 & \cellcolor{top1!45}0.664 \\
Text-IF~\cite{Text-IF}    & 7.386 & 16.798 & 6.638 & \cellcolor{top2}0.703 & \cellcolor{top2}0.523 & 2.850 & 0.340 & 0.502 & 15.484 & \cellcolor{top1!50}0.732 \\
SAGE~\cite{wu2025every} & 7.009 & 15.275 & 5.679 & 0.640 & 0.437 & 3.004 & 0.327 & 0.448 & 14.193 & \cellcolor{top1!39}0.577 \\
DAFusion~\cite{guan2026domain} & 7.081 & 13.840 & 5.409 & 0.686 & 0.512 & 2.890 & 0.357 & 0.500 & 15.092 & \cellcolor{top1!41}0.605 \\
\midrule
RoES (Ours)& \cellcolor{top2}7.461 & \cellcolor{top1}20.039 & \cellcolor{top1}7.091 & \cellcolor{top1}0.776 & \cellcolor{top1}0.561 & \cellcolor{top1}3.811 & \cellcolor{top1}0.367 & \cellcolor{top1}0.532 & 15.570 & \cellcolor{top1!100}0.986 \\

\midrule
\multicolumn{11}{c}{\textbf{TNO Dataset}} \\
\midrule
DenseFuse~\cite{li2019densefuse} & 6.468 & 8.824 & 3.395 & 0.639 & 0.366 & 2.254 & 0.250 & 0.463 & 15.392 & \cellcolor{top1!21}0.350 \\
FusionGAN~\cite{fusiongan} & 6.566 & 6.502 & 2.447 & 0.453 & 0.213 & 2.366 & 0.212 & 0.409 & 13.964 & \cellcolor{top1!0}0.086 \\
PMGI~\cite{PMGI} & 6.087 & 8.726 & 3.417 & 0.589 & 0.332 & 2.349 & 0.223 & 0.456 & 14.819 & \cellcolor{top1!17}0.292 \\
U2Fusion~\cite{u2fusion} & 6.562 & 11.787 & 4.863 & 0.619 & 0.382 & 2.009 & 0.255 & 0.513 & 15.726 & \cellcolor{top1!32}0.480 \\
RFN-Nest~\cite{ref7} & 7.071 & 5.944 & 2.704 & 0.612 & 0.289 & 2.275 & 0.242 & 0.460 & 15.062 & \cellcolor{top1!10}0.214 \\
SeAFusion~\cite{SeAFusion} & 7.186 & 12.573 & \cellcolor{top2}4.964 & 0.696 & 0.425 & 2.836 & 0.263 & 0.476 & 14.917 & \cellcolor{top1!41}0.595 \\
SwinFusion~\cite{swinfusion} & 6.985 & 12.630 & 4.876 & 0.730 & 0.442 & \cellcolor{top2}3.114 & 0.292 & 0.489 & 14.885 & \cellcolor{top1!47}0.661 \\
TarDAL~\cite{liu2022tardal} & 7.189 & 12.319 & 4.202 & 0.606 & 0.357 & 2.759 & 0.242 & 0.438 & 14.579 & \cellcolor{top1!31}0.472 \\
YDTR~\cite{ydtr} & 6.597 & 10.124 & 3.646 & 0.676 & 0.382 & 2.706 & 0.264 & 0.434 & \cellcolor{top1}16.229 & \cellcolor{top1!30}0.451 \\
\midrule
CDDFuse~\cite{zhao2023cddfuse} & 7.156 & 12.776 & 4.636 & 0.737 & 0.442 & 2.924 & 0.286 & 0.489 & 15.003 & \cellcolor{top1!45}0.640 \\
IRFS~\cite{IRFS} & 6.727 & 9.560 & 3.440 & 0.623 & 0.355 & 2.250 & 0.250 & 0.443 & 15.350 & \cellcolor{top1!22}0.353 \\
LRRNet~\cite{lrrnet} & 7.068 & 9.752 & 3.828 & 0.581 & 0.321 & 2.600 & 0.229 & 0.484 & 15.673 & \cellcolor{top1!23}0.365 \\
SegMiF~\cite{liu2023segmif} & 7.154 & 12.151 & 4.605 & 0.759 & 0.484 & 2.826 & 0.307 & \cellcolor{top2}0.526 & 15.344 & \cellcolor{top1!48}0.678 \\
EMMA~\cite{EMMA} & \cellcolor{top2}7.249 & 12.184 & 4.921 & 0.713 & 0.410 & 2.911 & 0.267 & 0.502 & 15.239 & \cellcolor{top1!42}0.599 \\
Text-IF~\cite{Text-IF} & \cellcolor{top1}7.255 & \cellcolor{top1}13.276 & \cellcolor{top1}5.018 & 0.728 & 0.454 & 2.923 & 0.276 & 0.513 & 15.490 & \cellcolor{top1!46}0.654 \\
SAGE~\cite{wu2025every} & 7.094 & 11.676 & 4.396 & 0.732 & 0.452 & 2.746 & 0.327 & 0.470 & 14.479 & \cellcolor{top1!44}0.635 \\
DAFusion~\cite{guan2026domain} & 6.829 & 10.252 & 3.946 & \cellcolor{top2}0.782 & \cellcolor{top2}0.493 & 2.640 & \cellcolor{top2}0.345 & 0.488 & 15.274 & \cellcolor{top1!45}0.636 \\
\midrule
RoES (Ours) & 7.207 & \cellcolor{top2}12.955 & 4.670 & \cellcolor{top1}0.928 & \cellcolor{top1}0.555 & \cellcolor{top1}3.741 & \cellcolor{top1}0.401 & \cellcolor{top1}0.530 & \cellcolor{top2}16.058 & \cellcolor{top1!100}0.950 \\
\bottomrule
\end{tabular}%
}
\label{tab:combined_evaluation}
\vspace{-6pt}
\end{table}

\subsection{Main Results}

\subsubsection{Quantitative Comparison}
Table~\ref{tab:combined_evaluation} reports cross-dataset results on M$^3$FD, RoadScene, and TNO. RoES obtains the best TOPSIS score on all three datasets, with clear margins over the second-ranked methods. It leads on SF, VIF, $Q_{abf}$, MI, FMI, and PSNR on M$^3$FD; ranks first on seven RoadScene metrics; and leads TNO on VIF, $Q_{abf}$, MI, FMI, $Q_{cb}$, and TOPSIS. These results show strong and balanced generalization across adverse-weather, road, and conventional fusion scenes.

RoES ranks first on MI, FMI, VIF, and $Q_{abf}$ throughout. The MI/FMI gains support RoUP's task-adaptive separation of shared low-frequency structure and complementary high-frequency information. The VIF/$Q_{abf}$ gains support the selective design: equivariant low-frequency modeling stabilizes global structure, while low-frequency-guided spectral refinement preserves details without rigidly constraining them. Together, these results demonstrate a favorable balance of information retention, structural fidelity, and detail enhancement.

\subsubsection{Qualitative Comparison}
Fig.~\ref{fig:fusion-vis-m3df} compares fusion results on M$^3$FD. Although all methods highlight thermal targets, their detail restoration and color fidelity differ markedly under severe smoke occlusion.

RoES preserves thermal targets, background consistency, and fine textures, whereas FusionGAN~\cite{fusiongan} and TarDAL~\cite{liu2022tardal} show color distortion or structural blur and SwinFusion~\cite{swinfusion} loses fine details. The results support frequency-selective equivariance for balancing thermal saliency and visible-detail recovery.

\subsubsection{Efficiency}
At $1024\times768$ resolution, RoES has 2.28M parameters, requires 1,989G FLOPs, and uses approximately 6.3GB of peak memory. Although RoES is not the lightest in FLOPs, its computational cost remains acceptable for practical deployment, and its peak memory usage fits within a single consumer GPU.

\begin{table}[thb!]
\centering
\caption{Ablation study on key components of RoES. }
\resizebox{\linewidth}{!}{%
\begin{tabular}{lccccccccc}
\toprule
Experiment & EN$\uparrow$ & SF$\uparrow$ & AG$\uparrow$ & VIF$\uparrow$ & $Q_{abf}$$\uparrow$ & MI$\uparrow$ & FMI$\uparrow$ & $Q_{cb}$$\uparrow$ & PSNR$\uparrow$ \\
\midrule
w/o Low-Eq          & 6.852 & \cellcolor{top2}17.161 & \cellcolor{top2} 5.422 & 0.908 & 0.615 & 3.578 & 0.390 & 0.481 & 17.427 \\
w/o Low-Eq with RotAug & 6.839 & 15.921 & 5.024 & 0.866 & 0.617 & 3.334 & 0.377 & 0.476 & 16.653 \\
w/o Low-Eq with $C_4$-TTA & 6.851 & \cellcolor{top1}17.171 & \cellcolor{top1}5.621 & 0.917 & 0.619 & 3.625 & 0.399 & 0.482 & 17.433 \\
w/ Fixed Wavelet       & 6.817 & 16.451 & 5.080 & 0.864 & 0.613 & 3.500 & 0.375 & 0.481 & 17.174 \\
w/o Dec. Sup.       & 6.848 & 16.734 & 5.168 & \cellcolor{top2}0.927 & 0.613 & 3.647 & 0.388 & 0.484 & 16.048 \\
w/o PSA             & 6.830 & 15.818 & 4.933 & 0.915 & \cellcolor{top2}0.627 & \cellcolor{top2}3.874 & \cellcolor{top2}0.401 & \cellcolor{top2}0.490 & \cellcolor{top2}17.749 \\
w/ Cartesian Attn.  & \cellcolor{top1}6.977 & 15.974 & 5.365 & 0.819 & 0.600 & 3.083 & 0.367 & 0.477 & 16.610 \\
\midrule
Baseline (RoES)     & \cellcolor{top2}6.858 & 15.932 &5.098 & \cellcolor{top1}0.941 & \cellcolor{top1}0.642 & \cellcolor{top1}4.135 & \cellcolor{top1}0.421 & \cellcolor{top1}0.495 & \cellcolor{top1}18.127 \\
\bottomrule
\end{tabular}%
}
\label{tab:ablation}
\vspace{-8pt}
\end{table}

\begin{table}[b!]
\centering
\caption{Quantitative evaluation of the downstream object detection task on the M$^3$FD dataset.}
\resizebox{\linewidth}{!}{%
\begin{tabular}{lccccccc}
\toprule
Method & Lamp & Car & Bus & Motor & Truck & People & mAP@0.5$\uparrow$ \\
\midrule
DenseFuse~\cite{li2019densefuse} & 0.464 & 0.820 & 0.710 & 0.310 & 0.483 & 0.668 & \cellcolor{top1!25}0.576 \\
FusionGAN~\cite{fusiongan} & 0.399 & 0.708 & 0.708 & 0.301 & 0.486 & 0.693 & \cellcolor{top1!13}0.549 \\
U2Fusion~\cite{u2fusion} & 0.433 & \cellcolor{top2}0.835 & 0.710 & 0.348 & 0.461 & 0.704 & \cellcolor{top1!27}0.582 \\
RFN-Nest~\cite{ref7} & \cellcolor{top2}0.497 & 0.830 & 0.702 & 0.385 & 0.438 & 0.671 & \cellcolor{top1!29}0.587 \\
SeAFusion~\cite{SeAFusion} & 0.398 & 0.826 & 0.719 & 0.340 & 0.471 & 0.694 & \cellcolor{top1!24}0.575 \\
SwinFusion~\cite{swinfusion} & 0.409 & 0.826 & 0.717 & 0.308 & \cellcolor{top2}0.516 & 0.686 & \cellcolor{top1!25}0.577 \\
TarDAL~\cite{liu2022tardal} & 0.451 & 0.824 & \cellcolor{top1}0.740 & 0.369 & 0.477 & \cellcolor{top2}0.727 & \cellcolor{top1!34}0.598 \\
YDTR~\cite{ydtr} & 0.338 & 0.820 & 0.682 & 0.263 & 0.415 & 0.604 & \cellcolor{top1!0}0.520 \\
\midrule
CDDFuse~\cite{zhao2023cddfuse} & 0.427 & 0.821 & 0.704 & 0.329 & 0.453 & 0.665 & \cellcolor{top1!20}0.566 \\
IRFS~\cite{IRFS} & 0.370 & 0.822 & 0.687 & 0.278 & 0.444 & 0.615 & \cellcolor{top1!7}0.536 \\
LRRNet~\cite{lrrnet} & 0.461 & 0.828 & 0.696 & 0.331 & 0.454 & 0.664 & \cellcolor{top1!23}0.572 \\
SegMiF~\cite{liu2023segmif} & 0.468 & 0.831 & \cellcolor{top2}0.727 & 0.365 & 0.438 & 0.684 & \cellcolor{top1!29}0.585 \\
EMMA~\cite{EMMA} & 0.462 & 0.827 & 0.647 & 0.392 & 0.473 & 0.682 & \cellcolor{top1!27}0.581 \\
Text-IF~\cite{Text-IF} & 0.398 & 0.814 & 0.681 & \cellcolor{top2}0.394 & 0.453 & 0.654 & \cellcolor{top1!20}0.566 \\
\midrule
RoES (Ours) & \cellcolor{top1}0.632 & \cellcolor{top1}0.849 & 0.659 & \cellcolor{top1}0.637 & \cellcolor{top1}0.565 & \cellcolor{top1}0.733 & \cellcolor{top1!100}0.679 \\
\bottomrule
\end{tabular}%
}
\label{tab:m3fd_detection}
\vspace{-8pt}
\end{table}

\subsection{Ablation Studies}
\textbf{Ablation Setup.} Table~\ref{tab:ablation} evaluates three components on M$^3$FD under the Sec.~4.1 protocol. \textit{w/o Low-Eq}, \textit{w/o Low-Eq with RotAug}, and \textit{w/o Low-Eq with $C_4$-TTA} compare built-in equivariance with augmentation and test-time averaging. \textit{w/ Fixed Wavelet} and \textit{w/o Dec. Sup.} test adaptive decomposition, while \textit{w/o PSA} removes low-frequency-guided spectral refinement. The \textit{w/ Cartesian Attn.} variant replaces PSA's polar projection with Cartesian cross-attention under otherwise identical settings.

\textbf{Ablation Analysis.} Removing low-frequency equivariance increases SF/AG by $1.229/0.324$ but reduces VIF, MI, $Q_{abf}$, and PSNR by $0.033$, $0.557$, $0.027$, and $0.700$, showing that equivariance improves fidelity rather than sharpness alone. Rotation augmentation does not recover the overall loss, whereas $C_4$-TTA partly recovers it. Fixed wavelets and removal of decomposition supervision also degrade performance; the latter decreases MI, FMI, and PSNR by $0.488$, $0.033$, and $2.079$. Removing PSA reduces VIF, MI, FMI, and PSNR by $0.026$, $0.261$, $0.020$, and $0.378$. The Cartesian variant trades minor EN/SF/AG gains for substantial fidelity loss, confirming the necessity of PSA's polar projection for accurate cross-frequency mapping. Thus, adaptive decomposition, low-frequency equivariance, and PSA make complementary contributions.

\begin{figure}[t!]
    \centering
    \resizebox{\linewidth}{!}{%
    \includegraphics[width=1\linewidth]{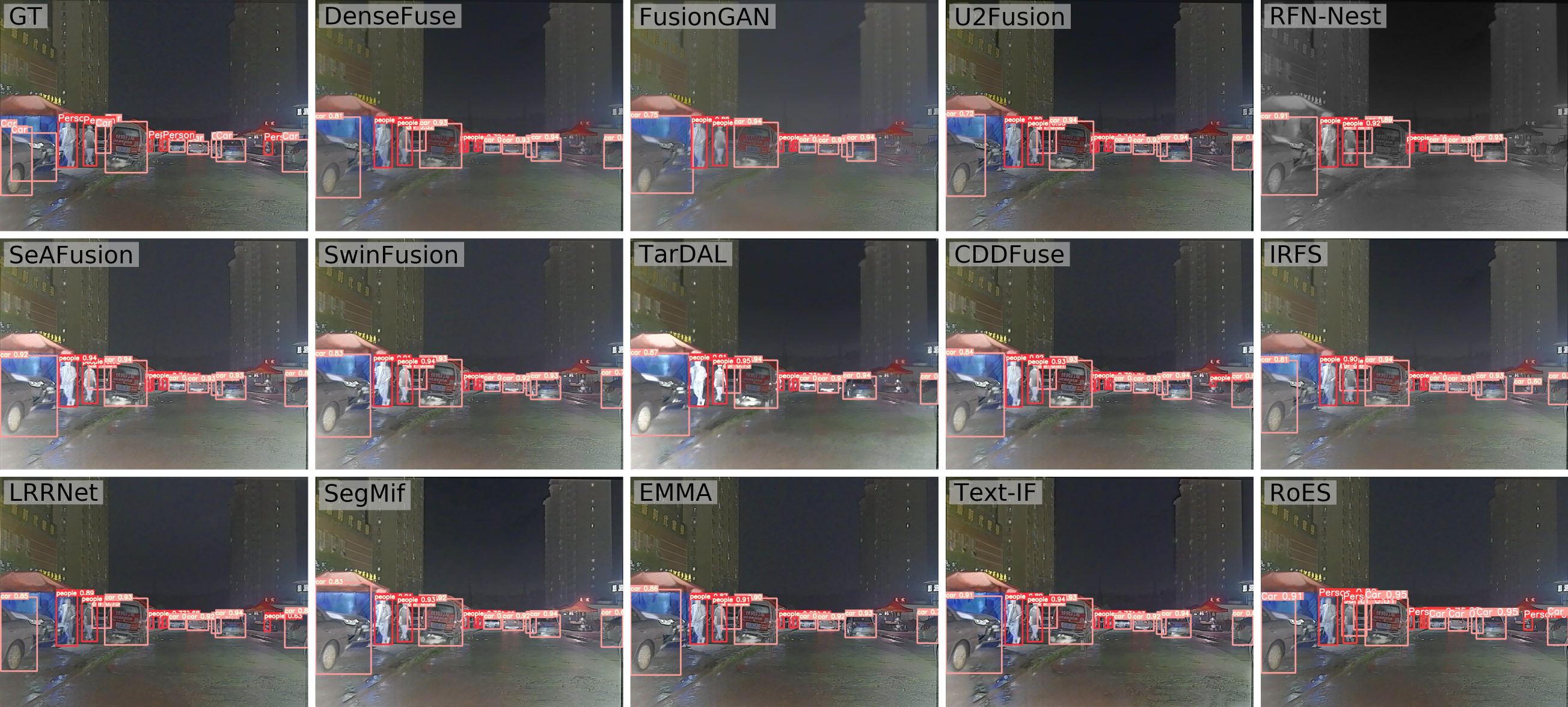}}
    \caption{Object detection results on the M$^3$FD dataset.}
    \label{fig:detection-vis}
    \vspace{-8pt}
\end{figure}
\subsection{Downstream Object Detection Task}

Following~\cite{mmf-pami-25}, we evaluate YOLOv5 v7.0 on M$^3$FD images fused by each method using the same detector architecture, with a separate detector trained for each fusion method under the same protocol. Table~\ref{tab:m3fd_detection} shows that RoES achieves 0.679 mAP@0.5, with strong results for Lamp (0.632), Motor (0.637), Car (0.849), and People (0.733). The fusion model is trained only on MSRS and transferred to M$^3$FD without fine-tuning, demonstrating cross-domain generalization across object scales.

Fig.~\ref{fig:detection-vis} further shows sharp contours, salient thermal targets, and precise boxes for small or distant objects. Most competing methods except LRRNet~\cite{lrrnet} miss or misalign pedestrians, supporting the benefit of stable low-frequency structure and preserved high-frequency texture.

\section{Conclusion}

We presented RoES, an infrared-visible fusion method that applies rotation equivariance selectively rather than uniformly. RoUP learns adaptive frequency decomposition; an equivariant Mamba branch models low-frequency structure; and a frequency-guided Dual-Fourier branch with polar spectral attention refines high-frequency details. Results across fusion benchmarks and downstream detection demonstrate state-of-the-art performance and support frequency-selective equivariance as an effective principle for robust multimodal fusion.

\begin{acks}
This work is partially supported by the National Natural Science Foundation of China (No. 62401632), the Hubei Provincial Natural Science Foundation (Nos. 2024AFB484 and 2026AFB758), the Natural Science Foundation of Wuhan (No. 2026040301020058), the Hubei Provincial Department of Education Key Projects (No. D20251003), the Open Fund of the Research Platforms, School of Computer Science, China University of Geosciences, Wuhan (No. PTLH2024-B-06), and the Fundamental Research Funds for the Central Universities of Zhongnan University of Economics and Law (No. 2722025BJ037).
\end{acks}

\bibliographystyle{ACM-Reference-Format}
\bibliography{ref}

\end{document}